\documentclass[10pt,twocolumn,letterpaper]{article}

\usepackage[pagenumbers]{cvpr} 

\definecolor{cvprblue}{rgb}{0.21,0.49,0.74}
\usepackage[pagebackref,breaklinks,colorlinks,allcolors=cvprblue]{hyperref}

\def\paperID{7620} 
\def\confName{CVPR}
\def\confYear{2025}

\definecolor{lightred}{rgb}{1, 0.25, 0.25}
\definecolor{darkgreen}{rgb}{0, 0.5, 0}
\definecolor{lightblue}{rgb}{0.18, 0.35, 0.9}

\title{Unlocking the Potential of Text-to-Image Diffusion with 
\textcolor{lightred}{P}\textcolor{darkgreen}{A}\textcolor{lightblue}{C}-Bayesian Theory}

\author{Eric Hanchen Jiang, Yasi Zhang, Zhi Zhang, Yixin Wan, Andrew Lizarraga, \\ Shufan Li, and Ying Nian Wu\\
University of California, Los Angeles\\
{\tt\small ericjiang0318@g.ucla.edu}
}

\usepackage{amsmath}      
\usepackage{amsthm}
\usepackage{amssymb}      
\usepackage{algorithm}    
\usepackage{algpseudocode}
\usepackage{graphicx}     
\usepackage{float}        
\usepackage{colortbl}
\usepackage{xcolor}

\newtheorem{definition}{Definition}

\begin{document}

\crefname{theorem}{Theorem}{Theorems}
\Crefname{theorem}{Theorem}{Theorems}

\crefname{lemma}{Lemma}{Lemmas}
\Crefname{lemma}{Lemma}{Lemmas}

\crefname{proposition}{Proposition}{Propositions}
\Crefname{proposition}{Proposition}{Propositions}

\crefname{corollary}{Corollary}{Corollaries}
\Crefname{corollary}{Corollary}{Corollaries}

\crefname{definition}{Definition}{Definitions}
\Crefname{definition}{Definition}{Definitions}

\crefname{assumption}{Assumption}{Assumptions}
\Crefname{assumption}{Assumption}{Assumptions}

\crefname{remark}{Remark}{Remarks}
\Crefname{remark}{Remark}{Remarks}

\twocolumn[
\vbox{
\maketitle
\centering
\includegraphics[width= 0.97\textwidth]{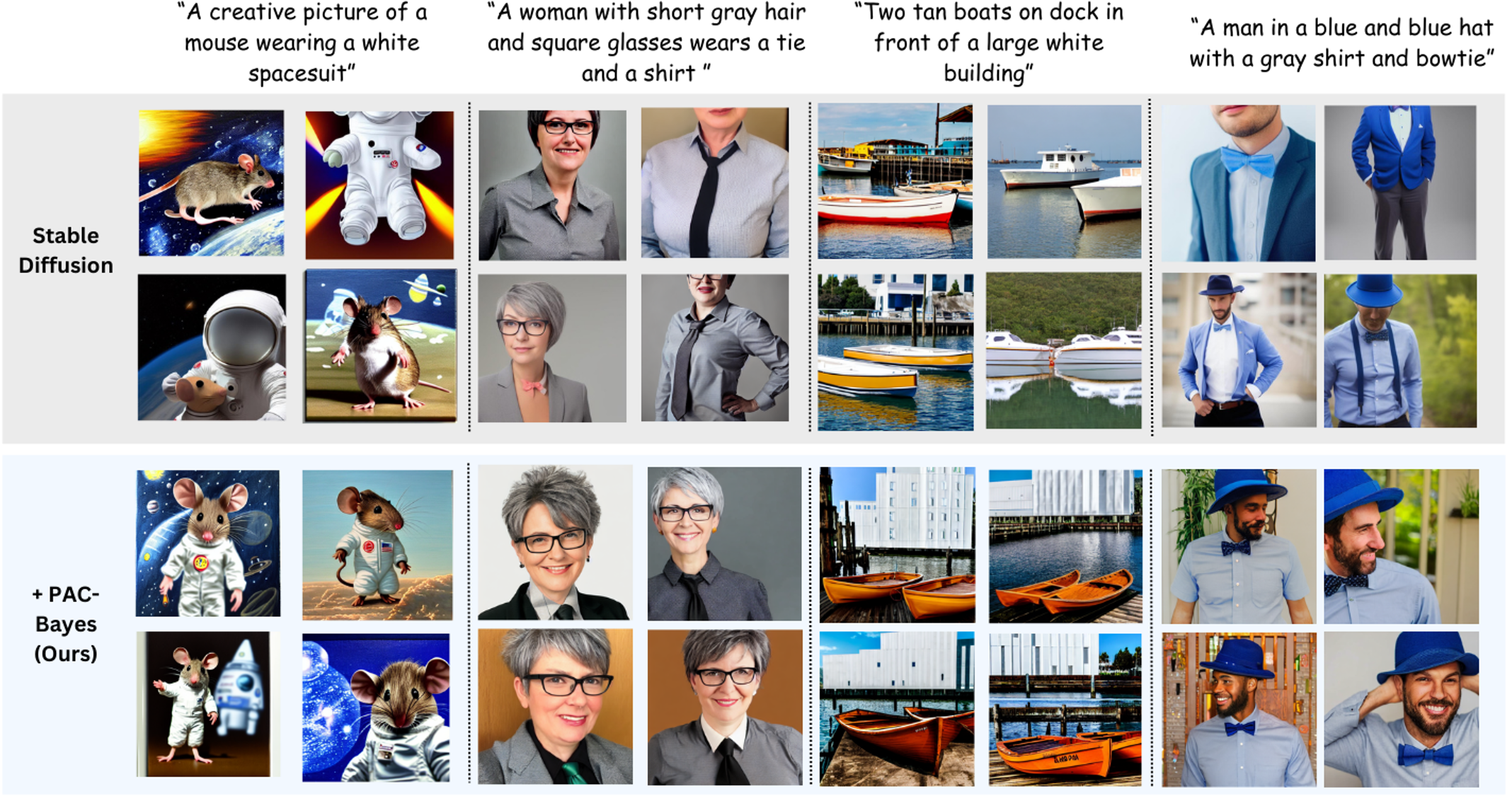}
\captionof{figure}{Diffusion models often struggle to accurately represent multiple objects in the input text. We identify the root causes of these challenges and introduce a training-free solution to address them by using PAC-Bayesian Theory. Here, we show some qualitative image results based on our model compared to the original Stable Diffusion model. For instance, when prompted to depict ``a mouse wearing a white spacesuit'', Stable Diffusion fails to separately associate individual modifiers with each component, therefore neglecting some of the descriptors (i.e. only generating a mouse or a spacesuit).}
\label{fig:intro_image}
\vspace{1em}
}]


\begin{abstract}
\vspace{-1em}


Text-to-image (T2I) diffusion models have revolutionized generative modeling by producing high-fidelity, diverse, and visually realistic images from textual prompts. Despite these advances, existing models struggle with complex prompts involving multiple objects and attributes, often misaligning modifiers with their corresponding nouns or neglecting certain elements. Recent attention-based methods have improved object inclusion and linguistic binding, but still face challenges such as attribute misbinding and a lack of robust generalization guarantees. Leveraging the PAC-Bayes framework, we propose a Bayesian approach that designs custom priors over attention distributions to enforce desirable properties, including divergence between objects, alignment between modifiers and their corresponding nouns, minimal attention to irrelevant tokens, and regularization for better generalization. Our approach treats the attention mechanism as an interpretable component, enabling fine-grained control and improved attribute-object alignment. We demonstrate the effectiveness of our method on standard benchmarks, achieving state-of-the-art results across multiple metrics. By integrating custom priors into the denoising process, our method enhances image quality and addresses long-standing challenges in T2I diffusion models, paving the way for more reliable and interpretable generative models.

\end{abstract}

\begin{figure}[t]
    \centering
    \includegraphics[width=0.475\textwidth]{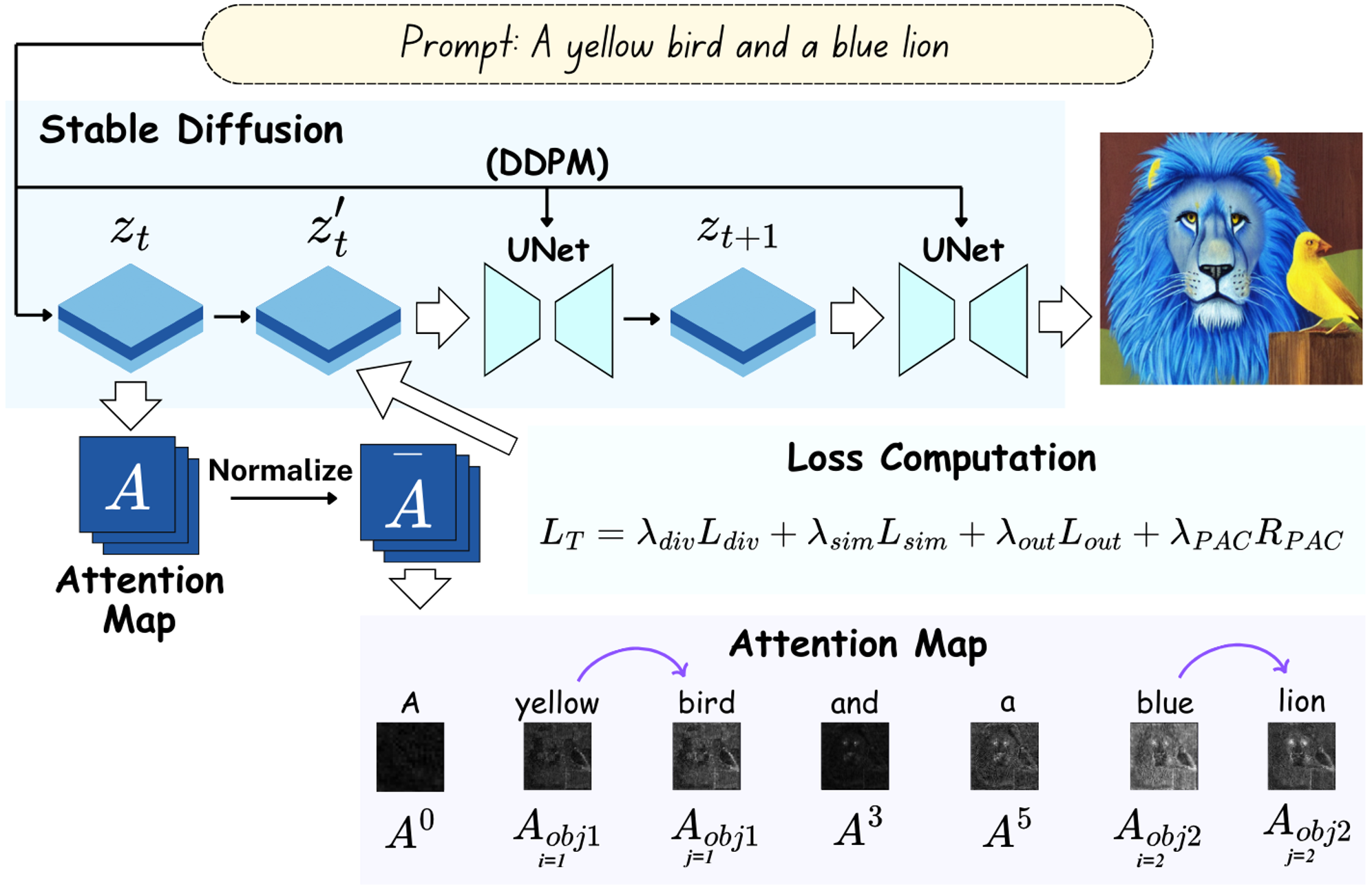} 
    \caption{\textbf{An overview of our workflow for optimizing the stable diffusion model.} It includes aggregation of attention maps, computation of object-centric attention loss, and updates to $z_t$.}
    \label{fig:flow_image}
\end{figure}

\section{Introduction} \label{sec
}

Recent text-to-image (T2I) diffusion models have achieved remarkable success in generating high-fidelity, diverse, and realistic images conditioned on textual descriptions~\cite{Rombach22, Nichol21, Saharia22, Ho20, Dhariwal21, Song21, Kingma21, Radford21, Esser21, Dosovitskiy21, Brock19, Goodfellow14, Reed2016generative}. Techniques such as classifier-free guidance~\cite{Ho2022classifier} and improved training strategies~\cite{Nichol2021improved} have further enhanced their performance. However, handling complex prompts with multiple objects and attributes remains challenging~\cite{Zhang24, Chefer23, Feng23, Liu22, Rassin23}. Diffusion models often misassociate attributes with incorrect nouns, leading to incorrect attribute bindings or omission of certain objects~\cite{Zhang24, Rassin23, Chefer23}. 

Challenges in attribute-object alignment are largely attributed to limitations within the models’ attention mechanisms, which operate as black boxes and lack transparency~\cite{Esser21, Rassin23, Feng23, Hertz23, Vaswani17}. Without explicit control over how attention is allocated, models struggle with tasks requiring precise modifier-object associations, resulting in semantic inaccuracies and compositional errors in the generated images~\cite{Hertz23, Avrahami2022blended, Brooks2023instructpix2pix}. Although various methods have attempted to guide attention, such as \emph{Prompt-to-Prompt} editing~\cite{Hertz23}, \emph{Composable Diffusion} models~\cite{Liu22}, and \emph{Training-Free Structured Diffusion Guidance}~\cite{Feng23}, issues like attribute misbinding and insufficient generalization persist~\cite{Zhang24, Chefer23}. 

In response to these limitations, we propose a novel Bayesian framework to enhance attention mechanisms within text-to-image diffusion models. Our approach leverages the PAC-Bayes framework~\cite{McAllester99} to allow users to design custom priors over attention distributions, enforcing desired properties such as divergence between different objects, similarity between modifiers and their respective nouns, minimized attention to irrelevant tokens, and a regularization component for robust generation. By minimizing the Kullback-Leibler divergence between learned attention distributions and custom-designed priors, our framework guides models towards improved attribute-object alignment and balanced attention allocation, effectively addressing limitations in previous approaches.

For instance, 
Moreover, issues such as object hallucination~\citep{Kim2022diffusionclip}, semantic confusion~\citep{Gal2022an}, and context misalignment~\citep{Li2023gligen} further exacerbate these challenges.

Our contributions can be summarized as follows:

\begin{itemize} \item We introduce a Bayesian framework that allows for custom priors over attention distributions, enhancing control and addressing the black-box nature of attention mechanisms. \item We formalize the problem within the PAC-Bayes framework~\cite{McAllester99}, providing theoretical generalization guarantees. \item Our method achieves state-of-the-art results on standard benchmarks~\cite{Chefer23, Feng23, Rassin23, Zhang24}, validating our approach's effectiveness in improving attribute binding and attention allocation. \end{itemize}

\section{Related Works}

\subsection{Stable Diffusion}

Stable Diffusion~\cite{Rombach22} represents a major advancement in text-to-image synthesis, utilizing a latent diffusion model to generate high-resolution images efficiently. By operating in a lower-dimensional latent space~\cite{yu2024latent} learned through an autoencoder~\cite{Esser21, Rombach21}, Stable Diffusion reduces computational demands compared to pixel-space diffusion models~\cite{Sohl-Dickstein2015deep, Ho20, Dhariwal21}. Cross-attention mechanisms~\cite{Vaswani17, Parmar2018image} enable Stable Diffusion to align textual and visual information, producing detailed images from textual prompts~\cite{Nichol21, Saharia22, Li2023}. Techniques like classifier-free guidance~\cite{Ho2022classifier} and improved training objectives~\cite{Nichol2021improved} have further enhanced generation quality and control~\cite{zhang2024flow, zhu2024think}. 

However, while these models excel at simple text-to-image tasks, they struggle with prompts involving multiple objects and attributes~\cite{Rassin23, Chefer23, Feng23}. Specifically, Stable Diffusion and similar models frequently misbind attributes to incorrect objects or overlook certain objects entirely, which limits their utility for tasks requiring precise modifier-object alignment~\cite{Zhang24, zhang2024objectconditioned,chang2024skews}. 

\subsection{Attention-Based Methods for Attribute-Object Alignment}


Several recent methods have been developed to improve attribute-object alignment by manipulating attention mechanisms. \citet{Hertz23} introduced \emph{Prompt-to-Prompt} editing, which enables users to control generation by adjusting cross-attention maps, and \citet{Liu22} proposed \emph{Composable Diffusion} models, where each object in the prompt can independently guide the output. Additionally, \citet{Feng23} presented \emph{Training-Free Structured Diffusion Guidance} to enhance compositionality. However, these methods often lack theoretical guarantees on model generalization and still face challenges with attribute misbinding~\cite{Zhang24, Chefer23}.

In related work, \citet{Chefer23} proposed the \emph{Attend-and-Excite} method, which amplifies attention toward specified objects to ensure their presence in the generated image. While effective for object inclusion, this approach can lead to unintended attribute leakage, where attributes from one object may spill over to another. Similarly, \emph{SynGen (SG)}~\cite{Rassin23} addressed attribute correspondence by leveraging linguistic binding to improve alignment within diffusion models. However, it also experiences challenges with nuanced attribute associations.

Another approach, \emph{Object-Conditioned Energy-Based Attention Map Alignment} (EBAMA)~\cite{Zhang24}, aligns attention maps between attributes and objects using an energy-based model, which improves semantic alignment. However, this method lacks flexibility for custom attention priors and relies heavily on predefined energy functions~\cite{Park23}. Other methods, such as \emph{A-STAR}~\cite{Agarwal23} and \emph{GLIGEN}~\cite{Li2023gligen}, incorporate external knowledge to guide attention but may increase model complexity and require additional resources.

\subsection{Bayesian Learning and the PAC-Bayes Framework}

The PAC-Bayes framework~\cite{McAllester99} offers theoretical tools for analyzing and bounding the generalization performance of stochastic learning algorithms. By integrating prior knowledge as a prior distribution over hypotheses, PAC-Bayes theory provides bounds that can inform model selection and regularization~\cite{zhang2024statistical}. Bayesian techniques have seen application across various machine learning domains, including neural networks and attention mechanisms~\cite{Ho2022classifier, Meng2021sdedit}, allowing for principled incorporation of uncertainty and prior knowledge. 

In diffusion models, Bayesian methods have shown promise for enhancing tasks like image restoration~\cite{Kawar2022denoising} and editing~\cite{Avrahami2022blended}. In this work, we apply the PAC-Bayes framework to formulate custom priors over attention distributions in diffusion models. This approach enables more interpretable and controlled attention mechanisms, resulting in improved attribute-object alignment and overall image coherence.

\section{Background}
\label{sec:background}

\subsection{Stable Diffusion Models}

Stable Diffusion Models (SD)~\cite{Rombach22}. SD first encodes an image \( x \) into the latent space using a pretrained encoder~\cite{Esser21}, i.e., \( z = E(x) \). Given a text prompt \( y \), SD optimizes the conditional denoising autoencoder \( \epsilon_\theta \) by minimizing the objective:
{\small
\begin{equation}
\mathcal{L}_\theta = \mathbb{E}_{t, \epsilon \sim \mathcal{N}(0,1), z \sim \mathcal{E}(x)} \left\| \epsilon - \epsilon_\theta(z_t, t, \phi(y)) \right\|^2,
\label{eq:sd_loss}
\end{equation}
}

where \( \phi \) is a frozen CLIP text encoder~\cite{Radford21}, \( z_t \) is a noised version of the latent \( z \), and the time step \( t \) is uniformly sampled from \( \{1, \dots, T\} \). During sampling, \( z_T \) is randomly sampled from standard Gaussian and denoised iteratively by the denoising autoencoder \( \epsilon_\theta \) from time \( T \) to \( 0 \). Finally, a decoder \( D \) reconstructs the image as \( \tilde{x} = D(z_0) \).

\subsection{Cross-Attention Mechanism}

In the cross-attention mechanism, \( K \) is the linear projection of \( W_y \), the CLIP-encoded text embeddings of text prompt \( y \). \( Q \) is the linear projection of the intermediate image representation parameterized by latent variables \( z \). Given a set of queries \( Q \) and keys \( K \), the (unnormalized) attention features and (softmax-normalized) scores between these two matrices are:

{\small
\begin{equation}
A = \frac{Q K^\top}{\sqrt{m}}, \quad \tilde{A} = \text{softmax} \left( \frac{Q K^\top}{\sqrt{m}} \right),
\label{equ:attn}
\end{equation}
}

where \( m \) is the feature dimension. We consider both attention features and scores for our modeling here, which we denote as \( A_s \) and \( \tilde{A}_s \) for token \( s \), respectively.

{\small
\begin{algorithm}[t]
\caption{A Single Denoising Step using PAC}
\label{alg:syngen_step}
\begin{algorithmic}[1]
\Statex \textbf{Input:} Text prompt $P$, latents $\mathbf{z}_t$, text embeddings $\mathbf{E}$, timestep $t$, iteration $i$, step size $\alpha_t$, number of intervention steps $K$, trained Stable Diffusion model $\text{SD}$, loss hyperparameters $\lambda_{\text{div}}$, $\lambda_{\text{sim}}$, $\lambda_{\text{out}}$, $\lambda_{\text{PAC}}$
\Statex \textbf{Output:} Updated latents $\mathbf{z}_t$
\State $\mathbf{z}_t \gets \mathbf{z}_t.\operatorname{detach}().\operatorname{requires\_grad\_}(True)$
\For{each latent $\mathbf{z}_t^{(n)}$ and text embedding $\mathbf{E}^{(n)}$}
    \State Reset gradients
    \State Compute noise prediction and attention maps: \par
    \hskip\algorithmicindent $(\boldsymbol{\epsilon}_\theta, \mathbf{A}_t) \gets \text{SD}(\mathbf{z}_t^{(n)}, P, t, \mathbf{E}^{(n)})$
    \State Aggregate attention maps: \par
    \hskip\algorithmicindent $\{\mathbf{A}_i\} \gets \operatorname{AggregateAttentionMaps}(\mathbf{A}_t)$
    \State Compute total loss: \par
    \hskip\algorithmicindent $\mathcal{L} = \lambda_{\text{div}} \mathcal{L}_{\text{div}} + \lambda_{\text{sim}} \mathcal{L}_{\text{sim}} + \lambda_{\text{out}} \mathcal{L}_{\text{out}} + \lambda_{\text{PAC}} \mathcal{R}_{\text{PAC}}$
    \If{$i < K$}
        \State Update latent: $\mathbf{z}_t^{(n)} \gets \mathbf{z}_t^{(n)} - \alpha_t \nabla_{\mathbf{z}_t^{(n)}} \mathcal{L}$
    \EndIf
\EndFor
\State \textbf{Return} $\mathbf{z}_t$
\end{algorithmic}
\end{algorithm}
}

\section{Method and Theoretical Foundations}
\label{sec:method}

In this section, we introduce our method designed to guide the attention mechanism of text-to-image diffusion models when processing prompts containing multiple objects. Our approach leverages the attention maps produced by the model to ensure balanced attention, correct attribute association, and improved generalization.

\subsection{Attention Map Distribution}

We focus on the extraction of nouns and modifiers from the text prompt and how the model can find their interactions. For example, given a prompt like \emph{"a yellow bird and a blue lion"}, we use a natural language parser to identify the nouns (\emph{"bird"}, \emph{"lion"}) and modifiers (\emph{"yellow"}, \emph{"blue"}), and determine the associations between them.

By analyzing the syntactic structure of the prompt, the parser can identify that \emph{"yellow"} modifies \emph{"bird"}, and not \emph{"lion"}. This information is crucial for guiding the attention mechanism so that the model correctly associates attributes with the corresponding objects in the generated image.

Our method utilizes spaCy’s transformer-based dependency parser \cite{spacy2} to project tokens in the prompt to their corresponding attention maps. By aligning the attention maps with the syntactic structure of the prompt, we can enforce constraints that encourage the model to attend appropriately to different parts of the image during generation.

\subsection{PAC-Bayes Framework}
We employ the PAC-Bayes framework~\cite{McAllester99}, which provides a theoretical foundation for understanding the generalization properties of learning algorithms.

\begin{definition}[PAC-Bayes Bound]
Given a prior distribution \( P \) over hypotheses \( h \in \mathcal{H} \), with probability at least \( 1 - \delta \), the following inequality holds:

{\small
\begin{equation}
\begin{aligned}
\mathbb{E}_{h \sim Q} [\text{Risk}(h)] &\leq \mathbb{E}_{h \sim Q} [\widehat{\text{Risk}}(h)] + \\
&\quad \sqrt{ \frac{ D_{\text{KL}}(Q \parallel P) + \ln \left( \dfrac{2 \sqrt{N}}{\delta} \right) }{2N} }
\end{aligned}
\label{eq:pac_bayes_bound}
\end{equation}
}

where \( Q \) is the posterior distribution over hypotheses, \( \text{Risk}(h) \) is the true risk, \( \widehat{\text{Risk}}(h) \) is the empirical risk, \( D_{\text{KL}}(Q \parallel P) \) is the KL divergence between \( Q \) and \( P \), \( N \) is the number of samples, and \( \delta \) is the confidence parameter (\( 0 < \delta < 1 \)).
\end{definition}

The PAC-Bayes framework allows us to derive generalization bounds that depend on the divergence between the learned model and a prior belief. By incorporating prior knowledge about how attention should be distributed in the model (e.g., modifiers should be associated with their corresponding nouns), we can guide the learning process in a principled way. By introducing the distributions over attentions, this method also helps to prevent overfitting and improves the generalization performance of the model. Notably, a wide range
of papers, from very theoretical to very computational, showed the use of PAC-Bayes bounds to improve the generalization performance of deep
networks \cite{zhou2018non, letarte2019dichotomize, tsuzuku2020normalized, clerico2023wide}. 


\begin{figure*}[th]
    \centering
    \includegraphics[width= \textwidth]{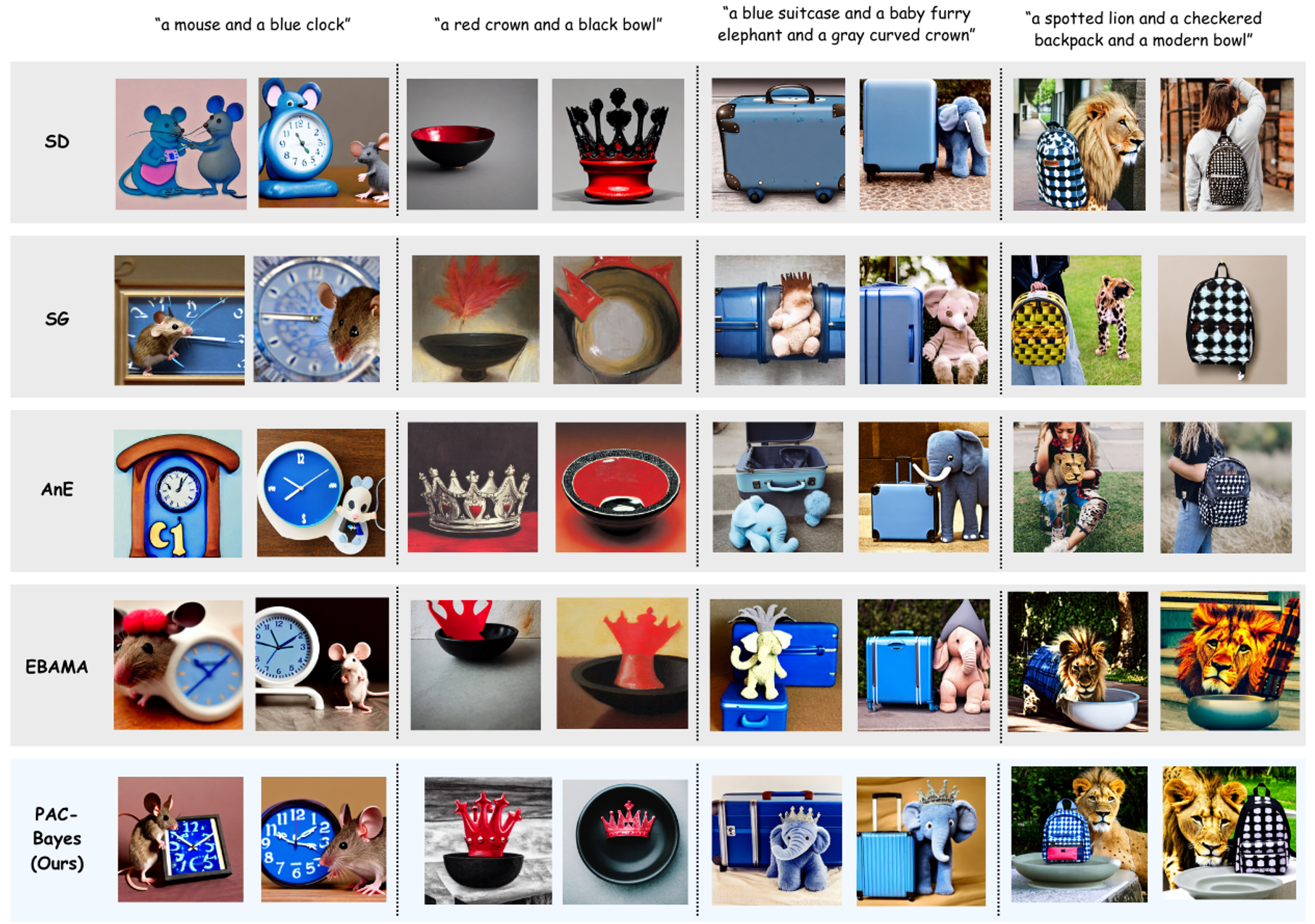}
    \caption{\textbf{Qualitative comparison on the AnE dataset (the left two columns) and the DVMP dataset (the right two columns).} We compared our model using the same prompt and random seed as SD \cite{Rombach21}, SG \cite{Rassin23}, AnE \cite{Chefer23}, and EMAMA \cite{Zhang24}, with each column sharing the same random seed.}
    \label{fig:ane_dvmp_image}
    \vspace{-1.2em}
\end{figure*}

\begin{figure*}[th]
    \centering
    \includegraphics[width=\textwidth]{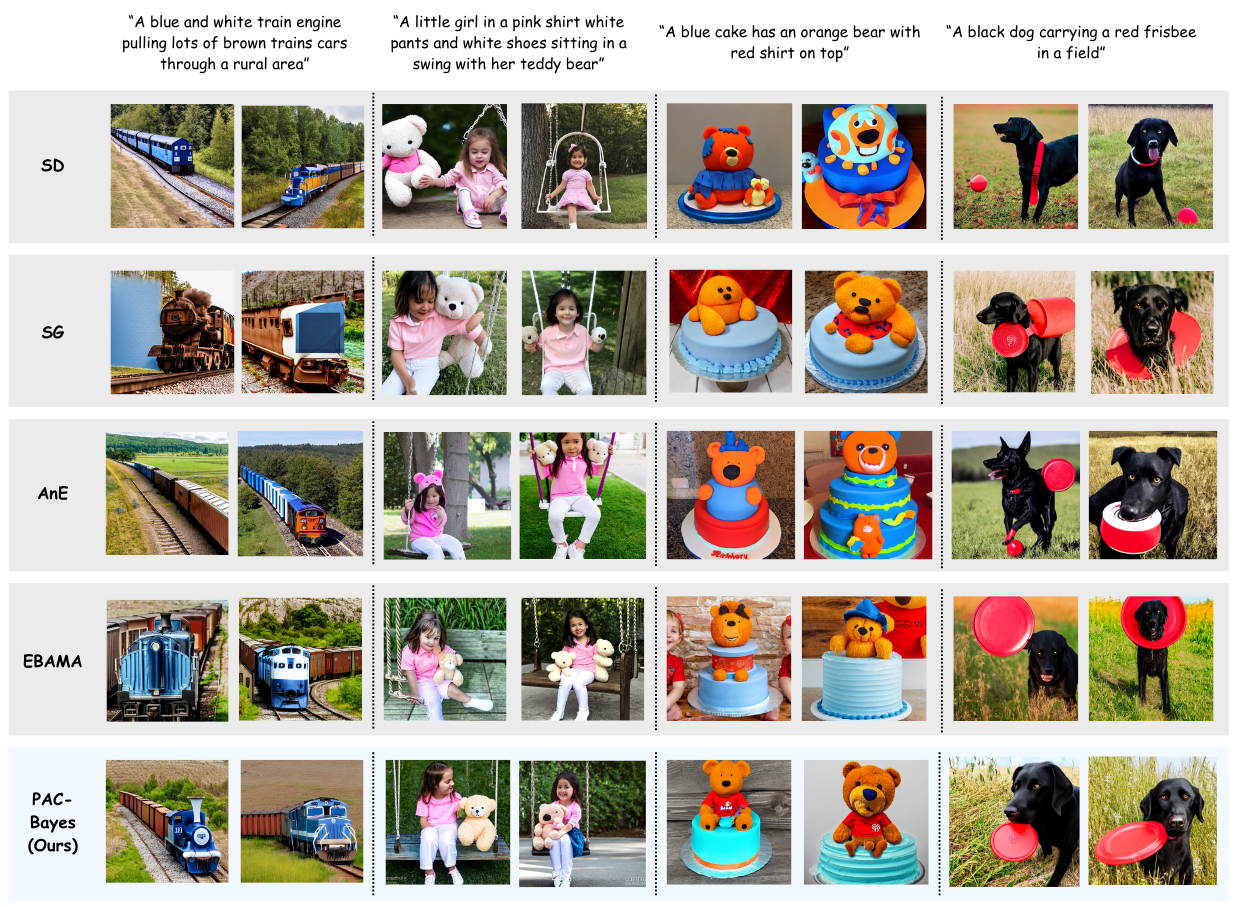}
    \caption{\textbf{Qualitative comparison on the ABC-6K dataset.} We compared our model using the same prompt and random seed as SD \cite{Rombach21}, SG \cite{Rassin23}, AnE \cite{Chefer23}, and EMAMA \cite{Zhang24}, with each column sharing the same random seed.}
    \label{fig:abc_image}
    \vspace{-0.8em}
\end{figure*}

\subsection{PAC-Bayes for Attention}

From the PAC-Bayes inequality in Equation~\eqref{eq:pac_bayes_bound}, PAC-Bayes theorem suggests optimizing the training loss along with a regularizer that incorporates a desired prior knowledge, providing a high confidence guarantee for the model's generalization. We present this novel PAC-Bayes framework for attention maps with distribution below with its detailed derivations found in Appendix. 
\subsubsection{Representation of Attention Maps}


We model attention maps 
as probability distributions
over a finite set \( \Omega \). For attention map $A_i$,  we let \( \Omega_i \) denote spatial indices within the image. 
Each entry \( A_i(j) \) represents a non-negative weight at spatial location $j$, satisfying:
\begin{equation}
A_i(j) \geq 0 \quad \forall j \in \Omega_i; \quad \sum_{j \in \Omega_i} A_i(j) = 1,
\label{eq:attention_distribution}
\end{equation} 



\subsubsection{Prior and Posterior Distributions}

Let the $U$ be the uniform distribution, as the prior.  The uniofrm is chosen to meet our desired criteria before learning, also see section~\ref{sec_regu} for the rationale.  


We define a multinomial-like structured  distribution \( A \) over attention maps that encode our desired properties.
Specifically, we factorize \( A \) as:

{ \footnotesize
\begin{equation}
\begin{aligned} &A(l) \propto \frac{1}{Z} \prod_{i \in \mathcal{A}_1} \left( A_i(l) \right)^{\alpha \Omega_i(l)} \prod_{j \in \mathcal{A}_2} \left( A_j(l) \right)^{\beta \Omega_j(l)} \prod_{k \in \mathcal{A}_3} \left( A_k(l) \right)^{\gamma \Omega_k(l)}
\label{eq:prior_factorization} \end{aligned}\end{equation}} 
where $Z$ is the normalization constant,  $\Omega_m(l)$ is an indicator function that equals 1 if $l\in \Omega_m$ for $m \in \{i, j, k\}$ and:



\begin{itemize}
    \item \( \mathcal A_1 \) corresponds to the divergence component, encouraging the attention maps of different objects to be separated.
    \item \( \mathcal A_2 \) corresponds to the similarity component, encouraging modifiers and their associated nouns to have similar attention maps.
    \item \( \mathcal A_3 \) corresponds to the outside component, discouraging attention to irrelevant tokens (e.g., articles like \emph{`a''}, \emph{the''}).
\end{itemize}

The exponents \( \alpha, \beta, \gamma\) are nonnegative hyperparameters controlling the influence of each component, satisfying \( \alpha + \beta + \gamma = 1 \).



The posterior \( A \) represents the learned attention distributions given the data.

\subsubsection{KL Divergence}

PAC-Bayes theory relies on the KL divergence between the posterior and the prior: 
We have 
{ \footnotesize
\begin{align}
&D_{\text{KL}}(A \parallel U) = \sum_{l} A(l) \log \left( \frac{A(l)}{U(l)} \right) \nonumber 
\end{align}
}
Substituting the factorized \( A \) from Equation~\eqref{eq:prior_factorization}, we have:

{ \footnotesize
\begin{align}
\label{kl_decomp}
D_{\text{KL}}(A \parallel U) &= \alpha \sum_{i \in \mathcal{A}_1} D_{\text{KL}}(A_i \parallel U) + \beta \sum_{j \in \mathcal{A}_2} D_{\text{KL}}(A_j \parallel U) \nonumber \\
&\quad +   \gamma \sum_{k \in \mathcal{A}_3}D_{\text{KL}}(A_k \parallel U)  + \text{C},
\end{align}
}
 where $C$ is a constant.

\subsection{Final Loss Function}
The final loss function combines the four components: divergence, similarity, outside, and regularization losses:


{\small
\begin{equation}
\mathcal{L}_{\text{total}} = \lambda_{\text{div}} \mathcal{L}_{\text{div}} + \lambda_{\text{sim}} \mathcal{L}_{\text{sim}} + \lambda_{\text{out}} \mathcal{L}_{\text{out}} + \lambda_{\text{PAC}} \mathcal{R}_{\text{PAC}},
\label{eq:total_loss_final}
\end{equation}
}
where \( \lambda_{\text{div}} = \alpha \), \( \lambda_{\text{sim}} = \beta \), \( \lambda_{\text{out}} = \gamma \), and \( \lambda_{\text{PAC}} = \eta \), corresponding to the exponents in  \eqref{eq:prior_factorization}.

By minimizing this loss according to the PAC-Bayes bound \cite{McAllester99}, we can control the generalization error of our model.

\subsubsection{Divergence Loss}

The divergence loss \(\mathcal{L}_{\text{div}}\) measures the symmetric KL divergence between the attention maps of all objects in a prompt. This loss encourages balanced attention allocation across a set of objects, \(\mathcal{O} = \{1, 2, \dots, n\}\), preventing any single object from dominating the attention map.

For each object \(k \in \mathcal{O}\), let \(A_{\text{obj}_k}\) represent its combined attention map. The divergence loss is defined as the average symmetric KL divergence over all unique object pairs in \(\mathcal{O}\):

{ \footnotesize
\begin{equation}
\mathcal{L}_{\text{div}} = - \frac{1}{|\mathcal{P}|} \sum_{(i, j) \in \mathcal{P}} \frac{1}{2} \left[ D_{\text{KL}}(A_{\text{obj}_i} \parallel A_{\text{obj}_j}) + D_{\text{KL}}(A_{\text{obj}_j} \parallel A_{\text{obj}_i}) \right]
\label{eq:divergence_loss}
\end{equation}
}
where \(\mathcal{P} = \{ (i, j) \mid i, j \in \mathcal{O}, i < j \}\) is the set of all unique object pairs in \(\mathcal{O}\).

This formulation enforces that the attention distributions across all objects are distinct and balanced, promoting clear separation in the attention space. By minimizing this divergence, the model is encouraged to differentiate each object effectively, ensuring that attention is allocated appropriately across the objects in the prompt.

\subsubsection{Similarity Loss}

The similarity loss \(\mathcal{L}_{\text{sim}}\) promotes coherence within each object by encouraging the attention maps of modifiers and nouns associated with the same object to be similar. For a given set of objects \(\mathcal{O} = \{1, 2, \dots, n\}\), each object \(k \in \mathcal{O}\) has associated modifiers and nouns.

Let \(\mathcal{M}_k\) denote the set of modifier indices and \(\mathcal{N}_k\) denote the set of noun indices for each object \(k\). Define \(\mathcal{P}_k = \{ (i, j) \mid i \in \mathcal{M}_k, \, j \in \mathcal{N}_k \}\) as the set of modifier-noun pairs for object \(k\). The total similarity loss \(\mathcal{L}_{\text{similarity}}\) is defined as the sum over all objects in \(\mathcal{O}\):

{\small
\begin{equation}
\mathcal{L}_{\text{sim}} = \sum_{k \in \mathcal{O}} \frac{1}{|\mathcal{P}_k|} \sum_{(i, j) \in \mathcal{P}_k} D_{\text{sym}}(A_i, A_j)
\label{eq:similarity_loss_combined}
\end{equation}
}
where \(D_{\text{sym}}(A_i, A_j)\) is the symmetric KL divergence between attention maps \(A_i\) and \(A_j\):

{\small
\begin{equation}
D_{\text{sym}}(A_i, A_j) = \frac{1}{2} \left[ D_{\text{KL}}(A_i \parallel A_j) + D_{\text{KL}}(A_j \parallel A_i) \right]
\label{eq:symmetric_kl}
\end{equation}
}

This loss encourages each modifier’s attention map to align closely with that of its associated noun, ensuring that attributes like colors or textures are correctly bound to their respective objects.











\begin{table*}[t]
    \centering
    \resizebox{\textwidth}{!}{
    \begin{tabular}{|c|ccc|ccc|ccc|}
        \hline
        \textbf{Method} & \multicolumn{3}{c|}{\textbf{Animal-Animal}} & \multicolumn{3}{c|}{\textbf{Animal-Object}} & \multicolumn{3}{c|}{\textbf{Object-Object}} \\
        \hline
         & \textbf{Full Sim.} & \textbf{Min. Sim.} & \textbf{T-C Sim.} & \textbf{Full Sim.} & \textbf{Min. Sim.} & \textbf{T-C Sim.} & \textbf{Full Sim.} & \textbf{Min. Sim.} & \textbf{T-C Sim.} \\
        \hline
        SD \cite{Rombach22} & 0.311 & 0.213 & 0.767 & 0.340 & 0.246 & 0.793 & 0.335 & 0.235 & 0.765 \\
        CD \cite{Liu22} & 0.284 & 0.232 & 0.692 & 0.336 & 0.252 & 0.769 & 0.349 & 0.265 & 0.759 \\
        StrD \cite{Feng23} & 0.306 & 0.210 & 0.761 & 0.336 & 0.242 & 0.781 & 0.332 & 0.234 & 0.762 \\
        EBCA \cite{Park23} & 0.291 & 0.215 & 0.722 & 0.317 & 0.229 & 0.732 & 0.321 & 0.231 & 0.726 \\
        AnE \cite{Chefer23} & 0.332 & 0.248 & 0.806 & 0.353 & 0.265 & 0.830 & 0.360 & 0.270 & 0.811 \\
        SG \cite{Rassin23} & 0.311 & 0.213 & 0.767 & 0.355 & 0.264 & 0.830 & 0.355 & 0.262 & 0.811 \\
        EBAMA \cite{Zhang24} & 0.340 & 0.256 & 0.817 & 0.362 & 0.270 & 0.851 & 0.366 & 0.274 & 0.836 \\
        D\&B \cite{Li2023} & 0.331 & 0.246 & 0.810 & -- & -- & -- & -- & -- & -- \\
        \hline
        \textbf{Ours} & \textbf{0.348} & \textbf{0.262} & \textbf{0.823} & \textbf{0.370} & \textbf{0.276} & \textbf{0.868} & \textbf{0.370} & \textbf{0.275} & \textbf{0.844} \\
        \hline
    \end{tabular}
    }
    \caption{ \textbf{ Comparison of Full Sim., Min. Sim., and T-C Sim. across different methods on the AnE dataset. CLIP Score based on the Attend and Excite Dataset.} Note that the performance of SG on Animal-Animal is degraded to SD, as the prompts do not contain any attribute-object pairs. The best results for each category is marked in bold numbers.}
    \label{tab:clip_scores}

\end{table*}

\subsubsection{Outside Loss}

The outside loss \(\mathcal{L}_{\text{out}}\) discourages the attention maps of object-related tokens from being similar to those of unrelated tokens, termed "outside" tokens. For a set of objects \(\mathcal{O} = \{1, 2, \dots, n\}\), define \(\mathcal{S} = \bigcup_{k \in \mathcal{O}} (\mathcal{M}_k \cup \mathcal{N}_k)\) as the set of all indices associated with object-related tokens. The set of outside tokens, \(\mathcal{O}_{\text{outside}}\), includes all other token indices not in \(\mathcal{S}\).

The outside loss is defined as:

{\small
\begin{equation}
\mathcal{L}_{\text{out}} = -\frac{1}{|\mathcal{S}|} \sum_{i \in \mathcal{S}} \left( \max_{j \in \mathcal{O}_{\text{out}}} D_{\text{sym}}(A_i, A_j) \right)
\label{eq:outside_loss}
\end{equation}
}

This loss penalizes high similarity between object-specific tokens and unrelated tokens, thereby encouraging the model to focus attention on relevant parts of the prompt and improving the clarity and accuracy of the generated images.





\subsubsection{PAC-Bayes Regularizer}
\label{sec_regu}

The PAC-Bayes regularizer \(\mathcal{R}_{\text{PAC}}\) applies principles from PAC-Bayes theory to regularize the attention distributions, promoting generalization and preventing overfitting. For a given set of samples, let \(N\) be the number of samples and \(\delta\) be the confidence parameter. The regularizer is defined as:

{\small
\begin{equation}
\mathcal{R}_{\text{PAC}} = -\sqrt{ \frac{ D_{\text{KL}}(A \parallel U) + \log \left( \frac{2 \sqrt{N}}{\delta} \right) }{2N} }
\label{eq:pac_bayes_regularizer}
\end{equation}
}
where \(D_{\text{KL}}(A \parallel U)\) represents the KL divergence between the learned attention distribution \(A\) and the uniform distribution \(U\) over the attention map dimensions.

The PAC-Bayes regularizer guides the attention map away from uniformity and 
encourages the attention map to focus on more informative patterns. This approach helps the model generalize by emphasizing meaningful structures in the data, while still avoiding overfitting.

\section{Workflow}


Our method integrates custom attention priors into the denoising process of text-to-image diffusion models, specifically focusing on Stable Diffusion~\cite{Rombach22}. We start by parsing the input prompt with a natural language parser to extract nouns and their associated modifiers, identifying relationships among them. In each denoising step, we capture the cross-attention maps for these tokens and apply custom loss functions guided by the PAC-Bayes framework~\cite{McAllester99}. These loss functions, based on the Kullback-Leibler divergence between the learned attention distributions and our designed priors, encourage divergence between different objects' attention maps, ensure similarity between modifiers and their nouns, minimize attention to irrelevant tokens, and include a regularization term to prevent overfitting. Together, these losses iteratively update the latent representations, refining the generated image to align more closely with the prompt and enhancing attribute binding accuracy, as outlined in Algorithm~\ref{alg:syngen_step}.

\begin{figure}[th]
    \centering
    \includegraphics[width=0.475\textwidth]{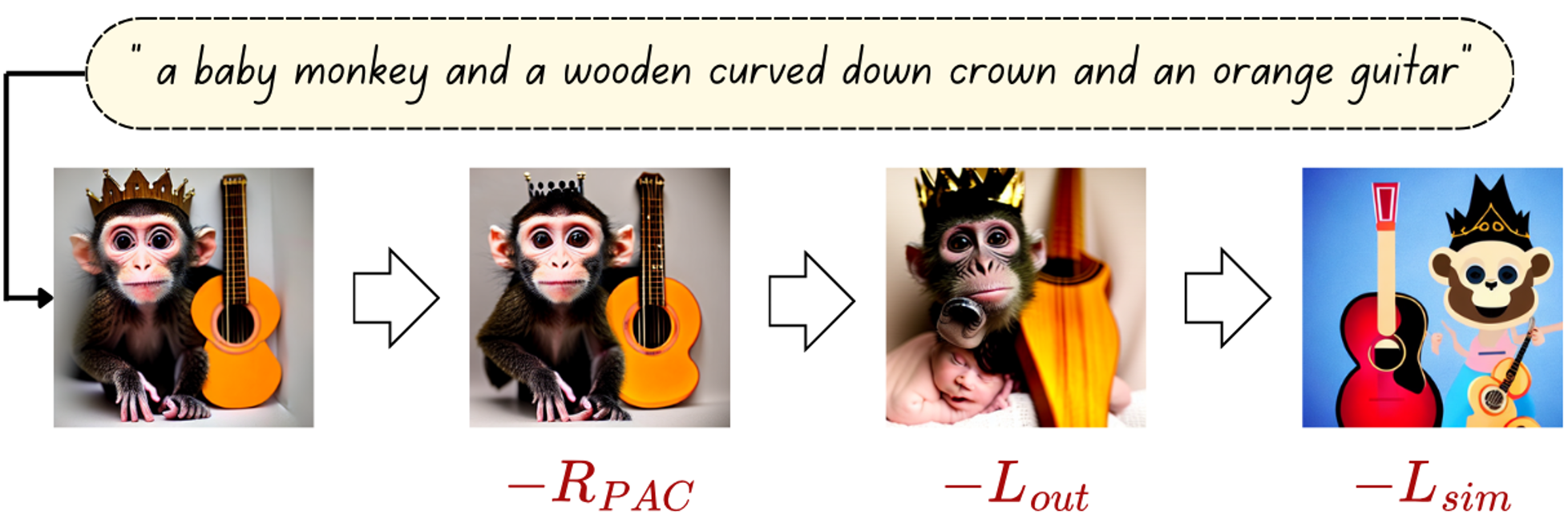}
    \caption{\textbf{Qualitative Results of the Ablation Study}. Each column illustrates the generated image for the prompt 'a baby monkey and a wooden curved crown and an orange guitar,' showing the effects of omitting specific loss components. From left to right: the full model output, the output without the PAC Regularizer, the output without the Outside Loss, and the output without the Similarity Loss. Each ablation demonstrates the impact on attribute-object alignment and overall image coherence, with attribute misbindings and inconsistencies appearing as each component is removed.}
    \label{fig:comparison_remove_loss}
\end{figure}

\section{Experiment}

\begin{figure}[th]
    \centering
    \includegraphics[width=0.475\textwidth]{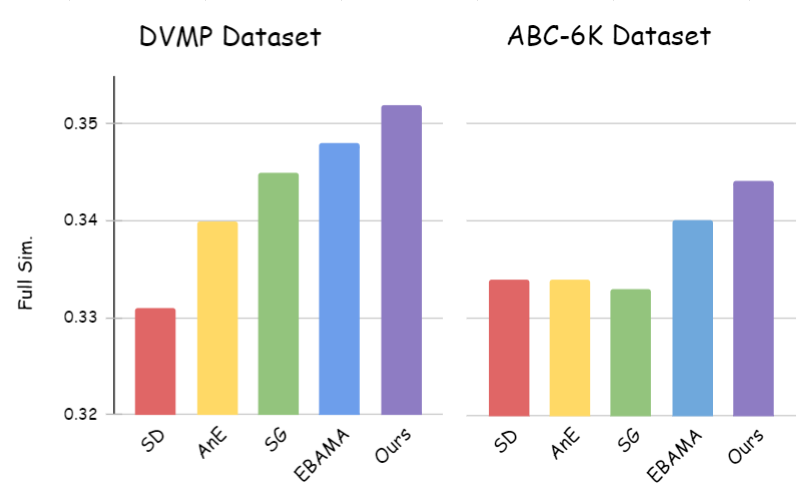}
    \caption{ \textbf{Full Sim. results on DVMP and ABC-6K datasets.} We randomly sample 200 prompts from each dataset and generate 4 images for each prompt.}
    \label{fig:chart_dvmp_abc_image}
\end{figure}

\begin{table}[h]
    \centering
    \begin{tabular}{|c|c c|}
        \hline
        \textbf{Model} & \textbf{Score} & \textbf{Change (\%)} \\
        \hline
        Ours & 0.344 & -- \\
        \hline
        - $\mathcal{R}_{PAC}$ & 0.338 & \textcolor{red!70!black}{-1.74\% $\downarrow$} \\
        \hline
        - $\mathcal{L}_{out}$ & 0.336 & \textcolor{red!70!black}{-0.59\% $\downarrow$} \\
        \hline
        - $\mathcal{L}_{sim}$ & 0.331 & \textcolor{red!70!black}{-1.48\% $\downarrow$} \\
        \hline
    \end{tabular}
    \caption{ \textbf{Impact of Removing Individual Loss Components on Full Similarity Scores} in the ABC Dataset. This table presents the Full Sim. scores for the complete model and each ablated variant, along with the corresponding percentage decrease in performance.}
    \label{tab:loss_ablation}
\end{table}

\textbf{Evaluation Datasets} \:
We evaluate the effectiveness of our proposed method on three benchmark datasets: Attend-and-Excite (AnE)~\cite{Chefer23}, ABC-6K, and DVMP~\cite{Rassin23}. The AnE dataset comprises three distinct benchmarks—Animal-Animal, Animal-Object, and Object-Object—each varying in complexity and incorporating combinations of colored animals and objects. Specifically, the prompt patterns include two unattributed animals, one unattributed animal paired with one attributed object, and two attributed objects, respectively. The DVMP dataset features a diverse set of objects, including daily objects, animals, and fruits, alongside multiple modifiers such as colors and textures, with each prompt containing more than three attribute descriptors. In contrast, the ABC-6K dataset consists of prompts with at least two color words modifying different objects. 

\textbf{Baselines} \:
Following the experimental setup of previous works~\cite{Chefer23,Zhang24}, we compare our method against several state-of-the-art models, including Stable Diffusion (SD)~\cite{Rombach22}, SyGen (SG)~\cite{Rassin23}, Attend-and-Excite (AnE)~\cite{Chefer23}, and EBAMA~\cite{Zhang24}. For each dataset, we generate images using identical prompts and random seeds across all methods to ensure a fair and consistent comparison.

\textbf{Evaluation Setup} \: To quantitatively assess the performance of our method, we employ three primary metrics based on CLIP similarity scores~\cite{Radford21}: Full Similarity (Full Sim.), Minimum Similarity (Min. Sim.), and Text-Caption Similarity (T-C Sim.). Full Sim. measures the cosine similarity between the CLIP embeddings of the entire text prompt and the generated image, providing an overall alignment score. Min. Sim. captures the lowest CLIP similarity score among all object-attribute pairs in the prompt and their corresponding regions in the image, highlighting the model's ability to accurately represent each element. T-C Sim. calculates the average CLIP similarity between the prompt and all captions generated by a pre-trained BLIP image-captioning model using the generated image as input, thereby evaluating the coherence and relevance of the generated content. For the DVMP and ABC-6K datasets, we utilize only the Full Sim. metric, measuring the overall alignment between the text prompt and the generated image. To ensure statistical robustness, we generate 64 images per prompt for the AnE dataset and 4 images per prompt for the DVMP and ABC-6K datasets, computing the average scores accordingly.

\textbf{Results and Analysis} \;
Figure~\ref{fig:ane_dvmp_image} showcases qualitative comparisons on the AnE and DVMP datasets, and Figure~\ref{fig:abc_image} showcases qualitative comparisons on the ABC datasets. Our method generates images with precise attribute associations and comprehensive inclusion of all specified objects, whereas baseline models frequently exhibit attribute misbinding or object neglect. From Table~\ref{tab:clip_scores}, we present a detailed comparison of CLIP similarity scores across different methods on the AnE dataset. Our method consistently achieves the highest scores across all categories—Animal-Animal, Animal-Object, and Object-Object—indicating superior attribute-object alignment and balanced attention allocation. Additionally, Figure~\ref{fig:chart_dvmp_abc_image} illustrates the Full Sim. results on the DVMP and ABC-6K datasets, where our approach outperforms baseline models, demonstrating its robustness across varying complexities.

\textbf{Ablation Study} To evaluate the contribution of each component in our loss function, we perform an ablation study on the ABC dataset, analyzing results both qualitatively and quantitatively. Figure~\ref{fig:comparison_remove_loss} illustrates the effects of removing individual components on a single prompt's text-to-image generation. Table~\ref{tab:loss_ablation} presents the Full Sim. scores when sequentially omitting the PAC-Bayes regularizer ($\mathcal{R}_{\text{PAC}}$), the outside loss ($\mathcal{L}_{\text{out}}$), and the similarity loss ($\mathcal{L}_{\text{sim}}$). Each removal leads to a decrease in performance, highlighting the essential role of each loss component in achieving optimal attribute-object alignment.

\section{Conclusion}

In our work, we introduced a Bayesian framework to enhance attention mechanisms in text-to-image diffusion models, addressing challenges in attribute-object alignment and balanced attention allocation. By leveraging PAC-Bayes principles, our method integrates custom attention priors to improve attribute binding, image quality, and generalization. Experiments demonstrated state-of-the-art performance across benchmarks, with ablations validating the impact of each loss component. This work offers a step toward more interpretable and controllable generative models, with potential for broader applications in text-to-image synthesis.


{
    \small
    \bibliographystyle{ieeenat_fullname}
    \bibliography{main}
}

\clearpage
\setcounter{page}{1}
\maketitlesupplementary

\section*{A. Limitations} 
\label{appendix:limitations}

While our method effectively guides attention in text-to-image diffusion models, it has limitations. It depends on the underlying Stable Diffusion model's capacity; if the base model struggles with complex concepts, our method may not overcome these issues. The framework requires identifiable object tokens in the text prompt; without them, it defaults to standard generation without our attention guidance. Incorporating our loss functions and latent updates adds computational overhead, potentially slowing generation times. Finally, the method relies on accurate natural language parsing; errors in parsing can lead to suboptimal attention guidance and affect image quality.

\section*{B. Societal Impact and Ethical Considerations} \label{appendix:ethics}

Our work advances text-to-image generation, with applications in art, design, and communication. However, it may be misused to create deceptive content or deepfakes, posing risks of misinformation and privacy violations. The models may also reflect biases from training data, potentially perpetuating stereotypes. Our method does not explicitly address these biases. Furthermore, generated images might resemble copyrighted material, raising intellectual property concerns. The computational demands contribute to energy consumption and environmental impact. We encourage responsible use and adherence to ethical guidelines to mitigate these issues.

\section*{C. Derivation of the PAC-Bayes Regularizer}
\label{appendix:pac_bayes}

In this section, we provide a detailed derivation of the PAC-Bayes regularizer used in our loss function. Starting from the PAC-Bayes bound~\cite{McAllester99}, for a prior distribution \( P \) and a posterior distribution \( Q \), the expected true risk \( \mathbb{E}_{h \sim Q} [\text{Risk}(h)] \) is bounded by:

{\footnotesize
\begin{equation}
\begin{aligned}
\mathbb{E}_{h \sim Q} [\text{Risk}(h)] &\leq \mathbb{E}_{h \sim Q} [\widehat{\text{Risk}}(h)] + \\
&\quad \sqrt{ \dfrac{ D_{\text{KL}}(Q \parallel P) + \ln \left( \dfrac{2 \sqrt{N}}{\delta} \right) }{2N} },
\end{aligned}
\label{eq:pac_bayes_bound_appendix}
\end{equation}
}

PAC-Bayes theory relies on the KL divergence between the posterior and the prior:

{\footnotesize
\begin{equation}
D_{\text{KL}}(A \parallel U) = \sum_{l} A(l) \log \left( \dfrac{A(l)}{U(l)} \right).
\label{eq:kl_divergence}
\end{equation}
}

Substituting the factorized \( A \) from Equation~\eqref{eq:prior_factorization}, we have:

{\footnotesize
\begin{align}
A(l) &= \dfrac{1}{Z} \prod_{i \in \mathcal{A}_1} \left( A_i(l) \right)^{\alpha\, \Omega_i(l)} \prod_{j \in \mathcal{A}_2} \left( A_j(l) \right)^{\beta\, \Omega_j(l)} \prod_{k \in \mathcal{A}_3} \left( A_k(l) \right)^{\gamma\, \Omega_k(l)}.
\label{eq:A_factorization}
\end{align}
}

Substituting this back into Equation~\eqref{eq:kl_divergence}, we get:

{ \footnotesize
\begin{align}
&D_{\text{KL}}(A \parallel U) 
= \sum_{l \in \Omega} A(l) \log \left( \frac{A(l)}{U(l)} \right) \nonumber \\
&= \sum_{l \in \Omega} A(l) \log \Bigg( 
    \frac{1}{U(l)} \cdot \frac{1}{Z} \prod_{i \in \mathcal{A}_1} 
          \left( A_i(l) \right)^{\alpha \mathbf{1}_{\Omega_i}(l)} 
    \cdot \frac{1}{Z} \prod_{j \in \mathcal{A}_2} 
          \left( A_j(l) \right)^{\beta \mathbf{1}_{\Omega_j}(l)} \nonumber \\
&\quad \cdot \frac{1}{Z} \prod_{k \in \mathcal{A}_3} 
          \left( A_k(l) \right)^{\gamma \mathbf{1}_{\Omega_k}(l)} 
    \Bigg).\\
&= \sum_{l \in \Omega} A(l) 
    \Bigg( \sum_{i \in \mathcal{A}_1} 
           \alpha \mathbf{1}_{\Omega_i}(l) \log \frac{A_i(l)}{U_i(l)} 
           + \sum_{j \in \mathcal{A}_2} 
           \beta \mathbf{1}_{\Omega_j}(l) \log \frac{A_j(l)}{U_j(l)} \nonumber \\
&\quad + \sum_{k \in \mathcal{A}_3} 
           \gamma \mathbf{1}_{\Omega_k}(l) \log \frac{A_k(l)}{U_k(l)} 
           - \log Z - C 
    \Bigg) \nonumber 
\end{align}
}

Since \( \Omega_m(l) \) is an indicator function that equals 1 if \( l \in \Omega_m \) and 0 otherwise, and \( A(l) \) is non-zero only when \( l \in \Omega_m \), we can simplify the sums:

{\footnotesize
\begin{align}
D_{\text{KL}}(A \parallel U) &= -\log Z \sum_{l} A(l) + \alpha \sum_{i \in \mathcal{A}_1} \sum_{l \in \Omega_i} A(l) \log A_i(l) \nonumber \\
&\quad + \beta \sum_{j \in \mathcal{A}_2} \sum_{l \in \Omega_j} A(l) \log A_j(l) + \gamma \sum_{k \in \mathcal{A}_3} \sum_{l \in \Omega_k} A(l) \log A_k(l) \nonumber \\
&\quad - \sum_{l} A(l) \log U(l).
\label{eq:kl_expanded}
\end{align}
}

Since \( \sum_{l} A(l) = 1 \) and \( \log Z \) and \( \log U(l) \) are constants, we can combine constants into a single term \( C \):

{\footnotesize
\begin{align}
D_{\text{KL}}(A \parallel U) &= \alpha \sum_{i \in \mathcal{A}_1} \sum_{l \in \Omega_i} A(l) \log A_i(l) + \beta \sum_{j \in \mathcal{A}_2} \sum_{l \in \Omega_j} A(l) \log A_j(l) \nonumber \\
&\quad + \gamma \sum_{k \in \mathcal{A}_3} \sum_{l \in \Omega_k} A(l) \log A_k(l) + C.
\label{eq:kl_simplified}
\end{align}
}

Recognizing that \( A_i(l) \) is the attention map for component \( i \) and \( A(l) \) aligns with \( A_i(l) \) when \( l \in \Omega_i \), we have:

{\footnotesize
\begin{align}
D_{\text{KL}}(A \parallel U) &= \alpha \sum_{i \in \mathcal{A}_1} D_{\text{KL}}(A_i \parallel U) + \beta \sum_{j \in \mathcal{A}_2} D_{\text{KL}}(A_j \parallel U) \nonumber \\
&\quad + \gamma \sum_{k \in \mathcal{A}_3} D_{\text{KL}}(A_k \parallel U) + C,
\label{eq:kl_decomposition}
\end{align}
}

where \( D_{\text{KL}}(A_i \parallel U) = \sum_{l \in \Omega_i} A(l) \log \left( \dfrac{A_i(l)}{U(l)} \right) \).

Therefore, optimizing \( D_{\text{KL}}(A \parallel U) \) is equivalent to optimizing the weighted sum of KL divergences of individual components.

By incorporating this into the PAC-Bayes bound in Equation~\eqref{eq:pac_bayes_bound_appendix}, we introduce the PAC-Bayes regularizer:

{\footnotesize
\begin{equation}
\mathcal{R}_{\text{PAC}} = -\sqrt{ \dfrac{ D_{\text{KL}}(A \parallel U) + \ln \left( \dfrac{2 \sqrt{N}}{\delta} \right) }{2N} }.
\label{eq:pac_bayes_regularizer}
\end{equation}
}

Piecing everything together, our final loss function becomes:

{\small
\begin{equation}
\mathcal{L}_{\text{total}} = \lambda_{\text{div}} \mathcal{L}_{\text{div}} + \lambda_{\text{sim}} \mathcal{L}_{\text{sim}} + \lambda_{\text{out}} \mathcal{L}_{\text{out}} + \lambda_{\text{PAC}} \mathcal{R}_{\text{PAC}},
\label{eq:total_loss}
\end{equation}
}

where \( \lambda_{\text{div}} = \alpha \), \( \lambda_{\text{sim}} = \beta \), \( \lambda_{\text{out}} = \gamma \), and \( \lambda_{\text{PAC}} = \eta \).

This formulation ensures that our loss function not only captures the empirical performance (through the divergence, similarity, and outside losses) but also incorporates theoretical guarantees on generalization provided by the PAC-Bayes framework.

\begin{figure}[t]
    \centering
    \includegraphics[width=0.475\textwidth]{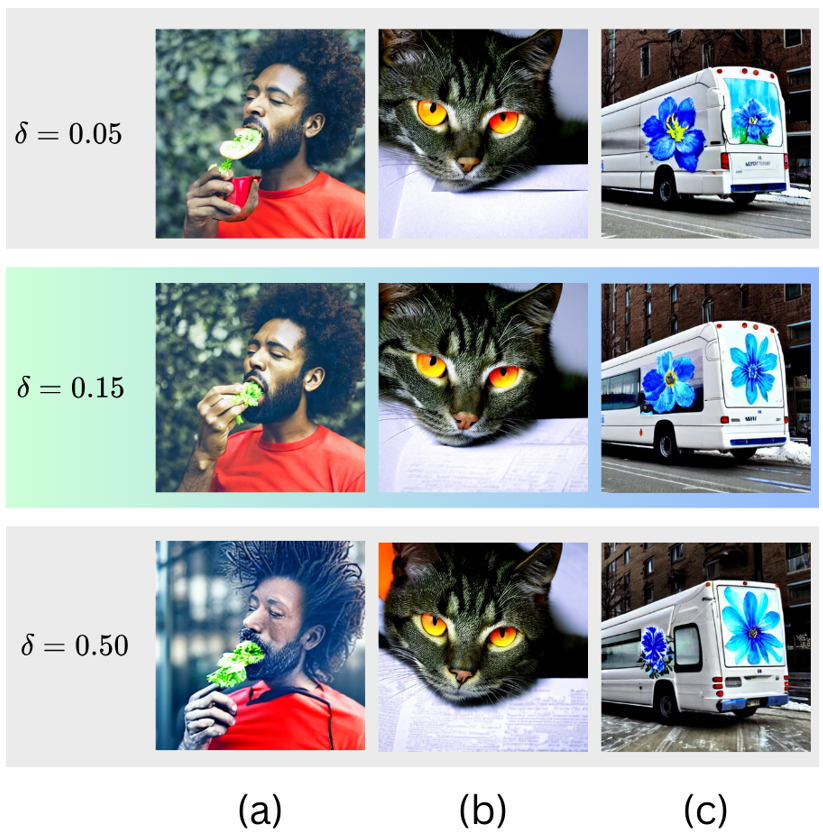} 
    \caption{\textbf{Ablation demonstration for PAC Regularizer's $\delta$} (a) a man in red shirt eating something that is green; (b) A green cat with orange eyes is laying over a paper; (c) A white bus with a painting of a blue flower on the front stopped on a street near a snow-covered sidewalk. }
    \label{fig:compare_delta}
\end{figure}

\section*{D. Implementation Details}
\label{appendix:implementation_details}

Our experiments were conducted on a system equipped with 4 NVIDIA A6000 GPUs capable of handling the computational demands of diffusion models and attention map computations. We utilized the official Stable Diffusion v1.4 model~\cite{Rombach22} and the CLIP ViT-L/14 text encoder~\cite{Radford21} for text embeddings.

For hyperparameters, we used a step size \( \alpha_t = 20 \), and the denoising steps was set to \( T = 50 \). The loss weights were set as \( \lambda_{\text{div}} = -1.25 \), \( \lambda_{\text{sim}} = 2.0 \), \( \lambda_{\text{out}} = 0.15 \), and \( \lambda_{\text{PAC}} = -0.15 \).

We employed the spaCy library~\cite{spacy2} with the \texttt{en\_core\_web\_trf} transformer model to parse the text prompts. The parser identifies nouns and their associated modifiers based on syntactic dependencies, enabling us to construct the sets \( \mathcal{M}_k \) and \( \mathcal{N}_k \). Attention maps were extracted from the cross-attention layers during the denoising process and normalized to ensure they are valid probability distributions over the spatial dimensions.

We used gradient descent to update the latent variables \( z_t \) based on the computed total loss. Gradients were computed with respect to \( z_t \), and updates were applied during the specified timesteps. To mitigate computational impact, we limited the latent updates to the first half of the denoising steps and optimized the implementation of attention map computations and loss evaluations.

\section*{E. Computational Efficiency}
\label{appendix:computational_efficiency}

\begin{table}[h!]
\centering
\begin{tabular}{|c|c|c|c|c|c|}
\hline
Method & SD & AnE & SG & EBAMA & Ours \\
\hline
Time (min) & 13.2 & 47.0 & 21.9 & 20.9 & 35.7 \\
\hline
\end{tabular}
\end{table}

Our method introduces additional computations due to the extraction and manipulation of attention maps and the application of custom loss functions during the denoising steps. As a result, the generation time per image increases compared to baseline models like Stable Diffusion. Despite this overhead, our method remains computationally feasible and can be optimized further; the trade-off in speed is justified by the significant improvements in image quality and attribute-object alignment.

\section*{F. Additional Ablation Experiments}
\label{appendix:additional_ablation}
\begin{figure}[t]
    \centering
    \includegraphics[width=0.475\textwidth]{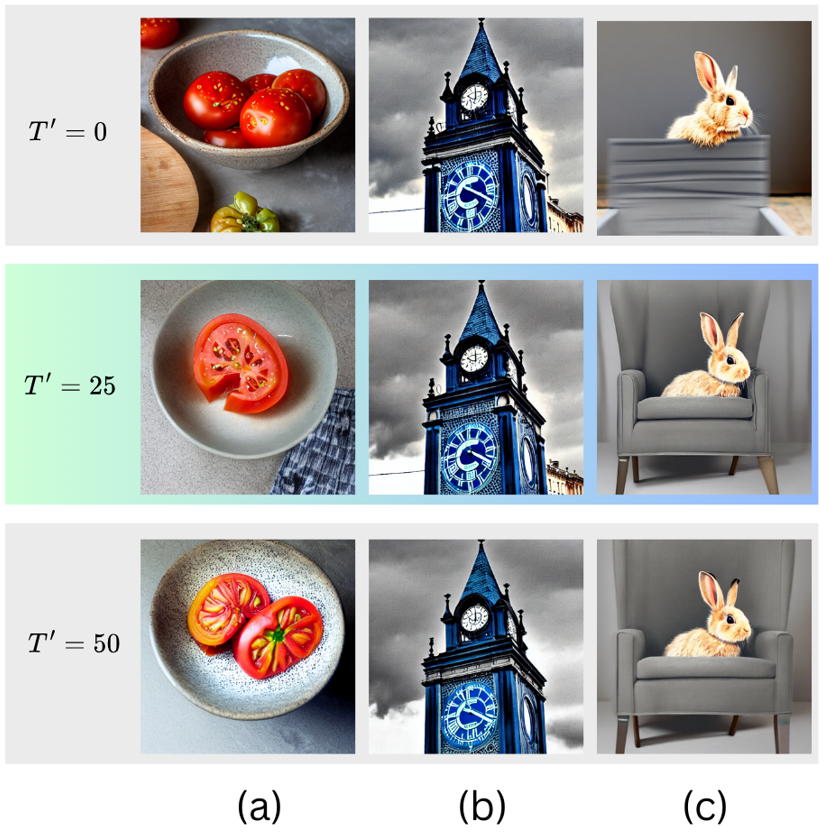} 
  \caption{\textbf{Ablation demonstration for time-steps $T'$} (a) a beige sliced tomato and a spotted bowl; (b) a blue clock tower is against the gray sky; (c) a rabbit and a gray chair. }
    \label{fig:compare_timesteps}
\end{figure}

\begin{figure}[h]
    \centering
    \includegraphics[width=0.475\textwidth]{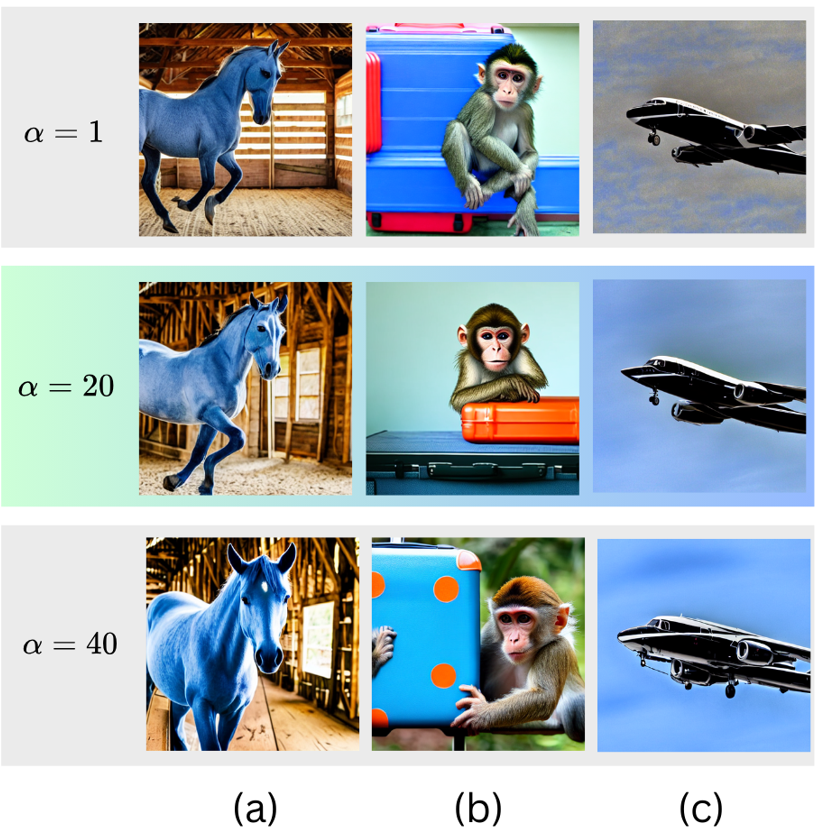} 
  \caption{\textbf{Ablation demonstration for step-size $\alpha$} (a) A young blue horse in a brown jacket running in a barn; (b) a monkey and a orange suitcase; (c) A black and white plane flying in a blue sky. }
    \label{fig:compare_alpha}
\end{figure}

In this section, we analyze the impact of varying key hyperparameters: the PAC-Bayes confidence parameter $\delta$, the number of updated timesteps $T'$, and the step size $\alpha$. Visual comparisons are provided in Figures~\ref{fig:compare_delta}, \ref{fig:compare_timesteps}, and \ref{fig:compare_alpha}, respectively, using the prompts from ABC-6K dataset.

\textbf{Confidence parameter $\delta$} We evaluate $\delta$ values of 0.05, 0.15 (ours), and 0.50 as shown in Figure~\ref{fig:compare_delta}. At $\delta = 0.05$, strong regularization leads to underfitting, resulting in poorly rendered objects: the green object in (a), the unnatural green cat in (b), and the faint blue flower in (c). Conversely, $\delta = 0.50$ weakens regularization, causing overfitting with dominant attributes overshadowing others: the red shirt in (a), excessive orange eyes in (b), and overly prominent blue flower in (c). Our choice of $\delta = 0.15$ balances regularization and flexibility, ensuring accurate attribute-object alignment and visual coherence.

\textbf{Timestep $T'$} We test $T' = 0$, $T' = 25$ (ours), and $T' = 50$ as depicted in Figure~\ref{fig:compare_timesteps}. With $T' = 0$, no updates lead to semantic misalignments, such as missing objects: the beige sliced tomato and spotted bowl in (a), incorrect clock tower color in (b), and absent gray chair in (c). Setting $T' = 25$ effectively binds attributes without introducing artifacts, accurately depicting all elements. However, $T' = 50$ results in distortions and visual artifacts, such as a distorted tomato in (a), unnatural textures in the clock tower in (b), and artifacts in the rabbit and chair in (c).

\textbf{Step size $\alpha$} We explore $\alpha = 1$, $\alpha = 20$ (ours), and $\alpha = 40$ in Figure~\ref{fig:compare_alpha}. A small step size of $\alpha = 1$ results in minimal updates, leading to insufficient attribute binding: the grayish blue horse and unclear brown jacket in (a), missing orange suitcase beside the monkey in (b), and incorrect black and white coloration of the plane in (c). Conversely, $\alpha = 40$ causes overemphasis and distortions: an overly saturated blue horse and dominant brown jacket in (a), overshadowed orange suitcase in (b), and a distorted black and white plane in (c). Our chosen step size of $\alpha = 20$ provides balanced updates, ensuring accurate and coherent image generation as seen in all examples.

\begin{figure*}[th]
    \centering
    \includegraphics[width=\textwidth]{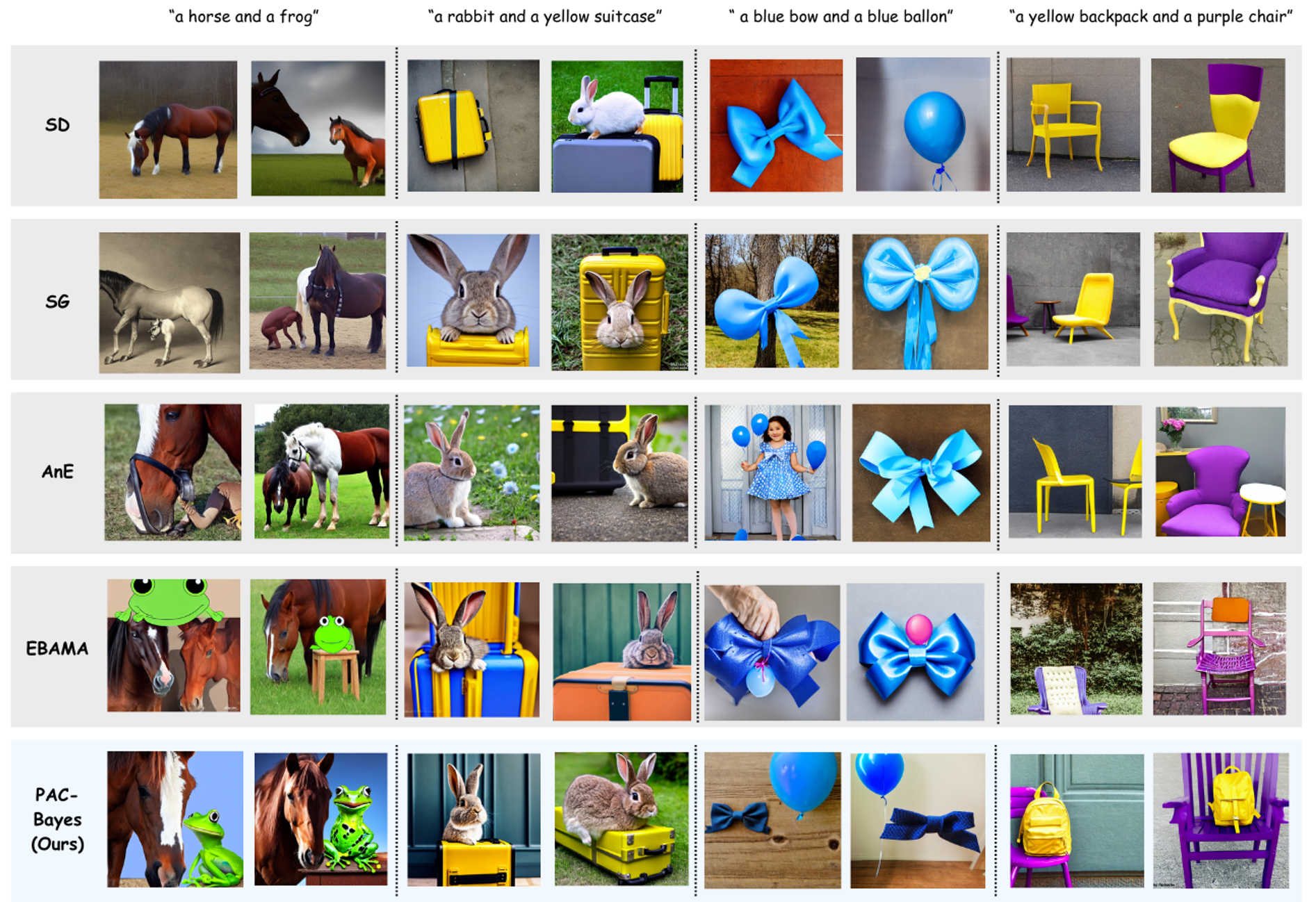}
    \caption{\textbf{Additional qualitative comparison on the AnE dataset.} We compared our model using the same prompt and random seed as SD \cite{Rombach21}, SG \cite{Rassin23}, AnE \cite{Chefer23}, and EMAMA \cite{Zhang24}, with each column sharing the same random seed.}
    \label{fig:body_image}
\end{figure*}

\begin{figure*}[th]
    \centering
    \includegraphics[width=\textwidth]{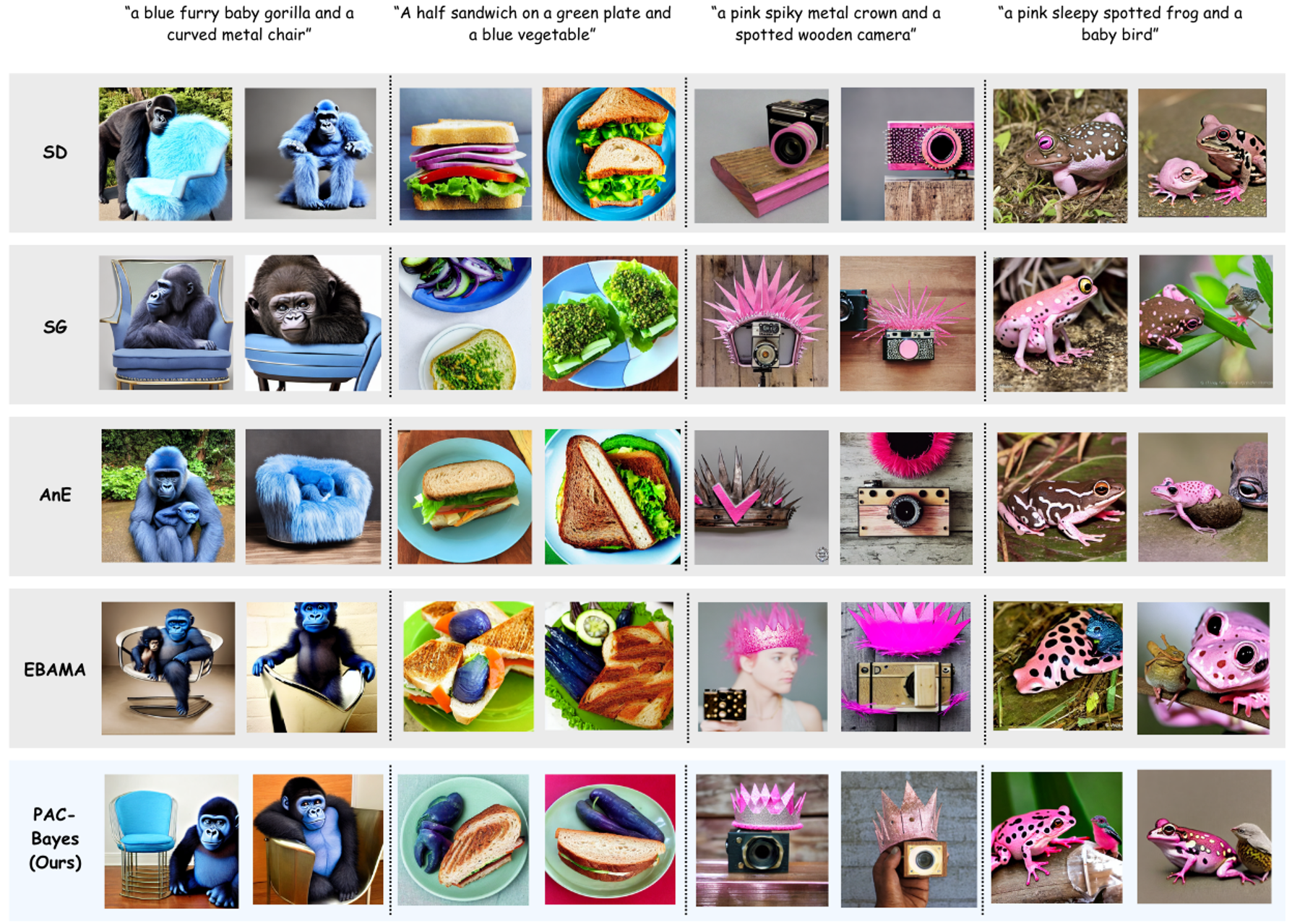}
    \caption{\textbf{Additional qualitative comparison on the DVMP dataset.} We compared our model using the same prompt and random seed as SD \cite{Rombach21}, SG \cite{Rassin23}, AnE \cite{Chefer23}, and EMAMA \cite{Zhang24}, with each column sharing the same random seed.}
    \label{fig:body_image}
\end{figure*}

\section*{G. Additional Visual Examples}
\label{appendix:visual_examples}

We present additional qualitative results to illustrate the effectiveness of our method. The examples demonstrate improved attribute binding and attention allocation compared to baseline models. The images generated using our approach show correct associations between modifiers and nouns, inclusion of all specified elements from the prompts, and balanced attention among multiple objects. These visual examples further validate the effectiveness of our Bayesian framework in guiding attention mechanisms and enhancing the quality of generated images, especially in complex prompts with multiple objects and attributes.

\end{document}